# Simulation and optimization of computed torque control 3 DOF RRR manipulator using MATLAB

Md Saad * [a], Sajjad Hussain [b]

[a] Department of Mechanical Engineering, Jamia Millia Islamia, New Delhi, India

[b] School of Arch, Tech and Engineering, University of Brighton UK.

**Abstract:** Robot manipulators have become a significant tool for production industries due to their advantages in high speed, accuracy, safety, and repeatability. This paper simulates and optimizes the design of a 3-DOF articulated robotic manipulator (RRR Configuration). The forward and inverse dynamic models are utilized. The trajectory is planned using the end effector's required initial position. A torque compute model is used to calculate the physical end effector's trajectory, position, and velocity. The MATLAB Simulink platform is used for all simulations of the RRR manipulator. With the aid of MATLAB, we primarily focused on manipulator control of the robot using a calculated torque control strategy to achieve the required position.

## 1. Introduction

In the automotive and production industries, robotic manipulators are widely used. A robotic manipulator with three degrees of freedom in the RRR configuration is considered. Nowadays, robots are in very high demand in the automation industry. They are responsible for industrial growth [1]. Robots are also used in the fields of medical science, assistance [2], nuclear power plants [3], virtual reality [4], and many fields in engineering. Robotics is a very growing field in engineering. There are three different types of control methods that may be identified: the first kind is traditional feedback control (PID and PD); the second type is adaptive control [5-7]; and the third type is iterative learning control (ILC) [8–10]. Some additional control methods, including model-based control, switching control, robust control [11], inverse dynamics control [12], and sliding mode control, may be reviewed in one way or another as combinations of those three basic types, or they may simply go by different names because of different emphasis when the three basic types are examined. There have been many academic papers written about modeling kinematics. Shi et al. [13] proposed a universal solution to the FK problem for a 6-DOF robot. Kumar et al. published an FK and IK solution for a virtual robot [14]. Cubero has suggested an IK for a robotic arm with a serial link [15]. In order to calculate the necessary joint angles for every desired arm position, Clothier et al. [16] have presented a geometric solution. The dynamic model is a crucial topic as we go toward control design [17]. Dynamic modeling incorporates forces and torques that are applied to the robot [18]. Numerous techniques have been developed by researchers to compute dynamics. Euler-Lagrange formulations and Newton-Euler are often used to model dynamics. For controller design, a mathematical model of the robotic arm is required. The kinematic model, as mentioned above, and the dynamic model are the two types of models utilized in robot modeling. To control the robotic manipulator precisely, even at higher speeds, a control strategy must be effectively established. Robot dynamics, payload, and operating environment present the greatest challenges in the development of the control system.

## 2. System modelling

A 3 DOF robotic manipulator with RRR configuration is used in this research work, having three revolute join and three links, which is shown in Fig 1. The robotic manipulator's link specification is specified in Table 1.

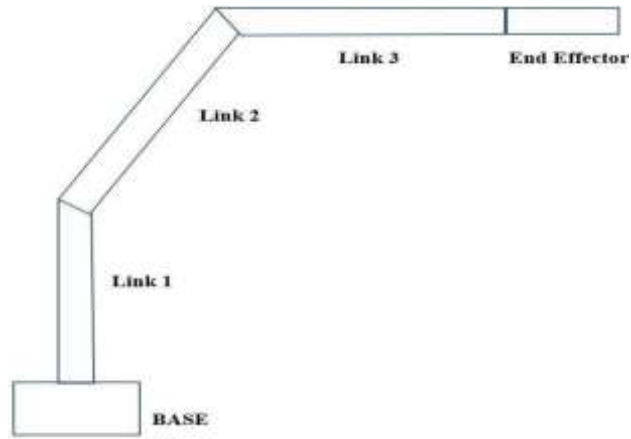

Fig 1 Schematic diagram of manipulator

|        | Length |
|--------|--------|
| Base   | 60 mm  |
| Link 1 | 150 mm |
| Link 2 | 300 mm |
| Link 3 | 200 mm |
| Link 4 | 50 mm  |

Table 1 Dimension of Manipulator

The manipulator is modeled and simulated in MATLAB Simulink using inverse and forward dynamic blocks from the robotics toolbox. To calculate the torque of the manipulator, the input trajectory must first provide the inverse dynamic model with information on position, velocity, and acceleration. Fig. 2 displays the input trajectory.

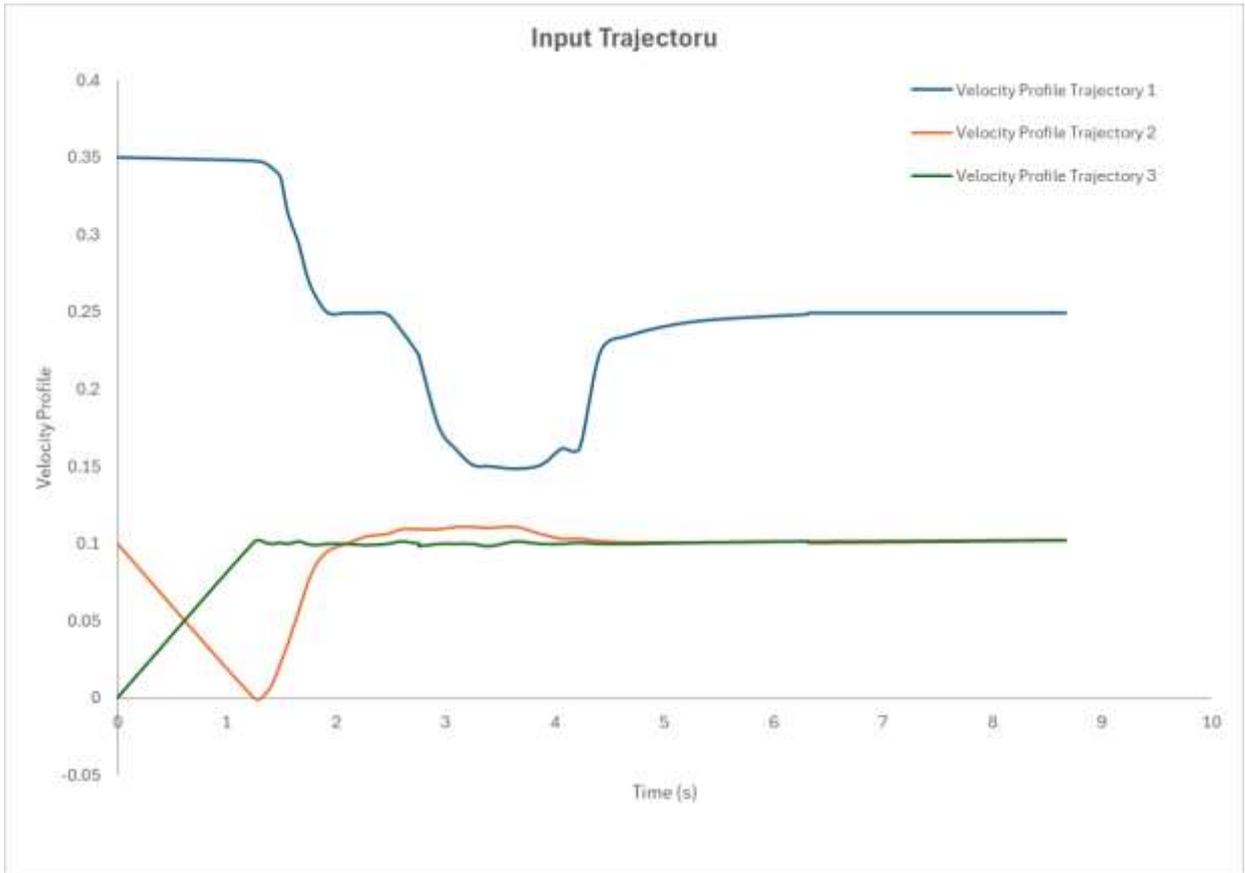

Fig 3 Input Trajectory

After the computed torque forward dynamic model is used to compute the actual position of the robotic manipulator and joint of robots. The CAD model of the robotic man is generated in MATLAB Simulink multibody which is shown in Fig 3.

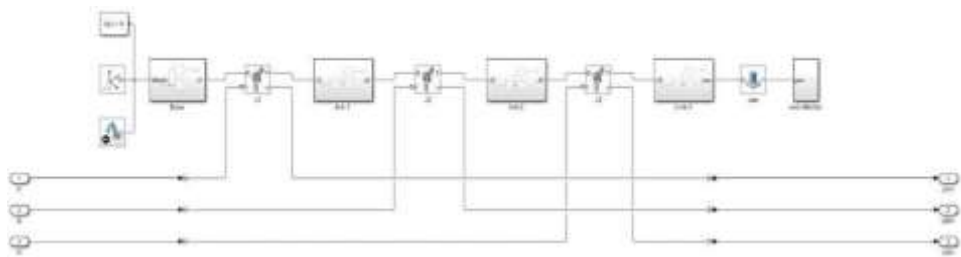

Fig 3 CAD in MATLAB

## 3. Dynamic model formulation

The relation between the joint torques generated by the actuators and the arm's location, velocity, and acceleration over time is examined by the robot's dynamical analysis. Precise control may be difficult due to the robot manipulators' complex and nonlinear dynamics. The dynamic equations of robot manipulators are often modeled by coupled non-linear differential equations derived from Lagrangians [21]:

$$Q = M(q)q'' + C(q, q') + G(q)$$

Where $M(q)q''$ is the inertia matrix, $C(q, q')$ is the centripetal matrix, $G(q)$ is the gravity vector, and Q is the control input torque. The joint variable is an n-vector containing the joint angles for revolute joints.

## 4. Computed Torque Control

Robot manipulators that are time-varying, dynamically coupled, and highly nonlinear are frequently used in industrial applications. The advantage of the dynamic model of the manipulator is that it can calculate the torque and force required to finish a typical work cycle and provides crucial information for the design of actuators, links, joints, motors, and control systems The manipulator's dynamic behavior offers a relationship between the joint actuator torques and the motion of the links, which is useful for simulation and the creation of control algorithms. CTC is a fundamental nonlinear control method used to eliminate the system's nonlinear behavior. The feedback loop for the control design will be decided by the dynamics that are described. When a precise dynamic model is present, CTC performs well and yields favorable performance metrics. Though not completely impossible, it is practically very challenging to create an accurate dynamic model. Furthermore, when a large amount of cargo is taken up, the dynamics of a robotic manipulator alter dramatically [25]. These modifications lead to a performance degradation of the manipulator's trajectory tracking. Numerous researchers have proposed cutting-edge CTC techniques to address this degradation, such as model-based fuzzy controllers to provide the necessary control and fuzzy switching controls to enhance closed-loop system performance. [26]. To estimate nonlinear dynamics, Soltani et al. presented fuzzy CTC in [27]. For robotic manipulator systems, computed torque control is an efficient motion control technique. Because it is simple to comprehend and performs well, computed torque control is noteworthy. Computed Torque Control is a technique for linearizing and decoupling the dynamics of robotic manipulator systems such that each joint motion can be independently controlled by more sophisticated linear control techniques. Calculated torque is a special application of feedback linearization of nonlinear systems that has gained acceptance in modern systems theory. The inversion of the robot dynamics is necessary for the computed-torque control, which is sometimes referred to as inverse dynamics control because it generates the joint acceleration vector after computing torque.

## 5. Result and Analysis

MATLAB/Simulink is used to run the simulation. Simulink's model is provided in Fig.

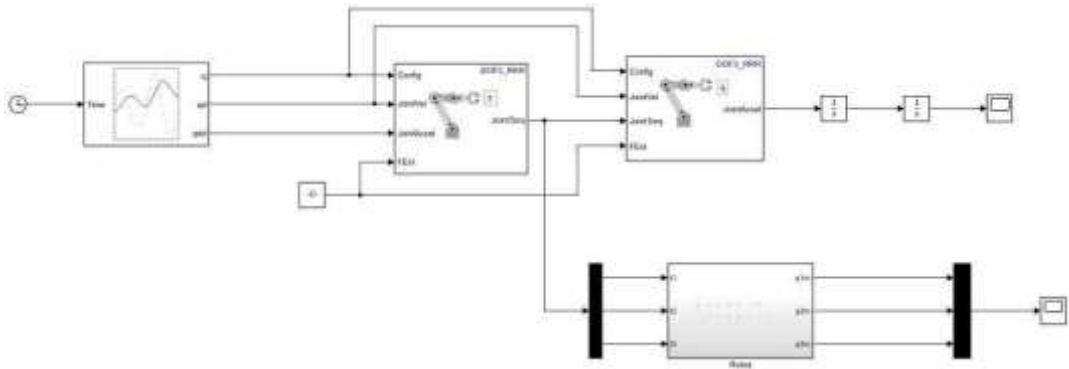

Fig-7 MATLAB Simulink Model

A robotic manipulator that has three degrees of freedom and three revolute joints is used. As seen in Fig. 3, the trajectory creation block in MATLAB Simulink is used to create the input trajectory for this simulation model. The end effector location and joint torque are calculated using an inverse and forward dynamic model. Figure 8-11 displays the simulations' outcome.

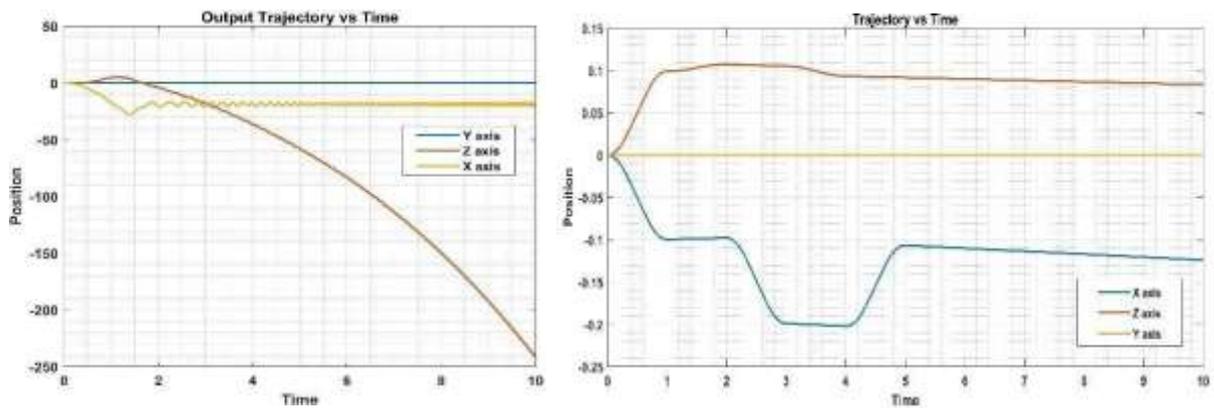

Fig 8 Output Position from Dynamic Model

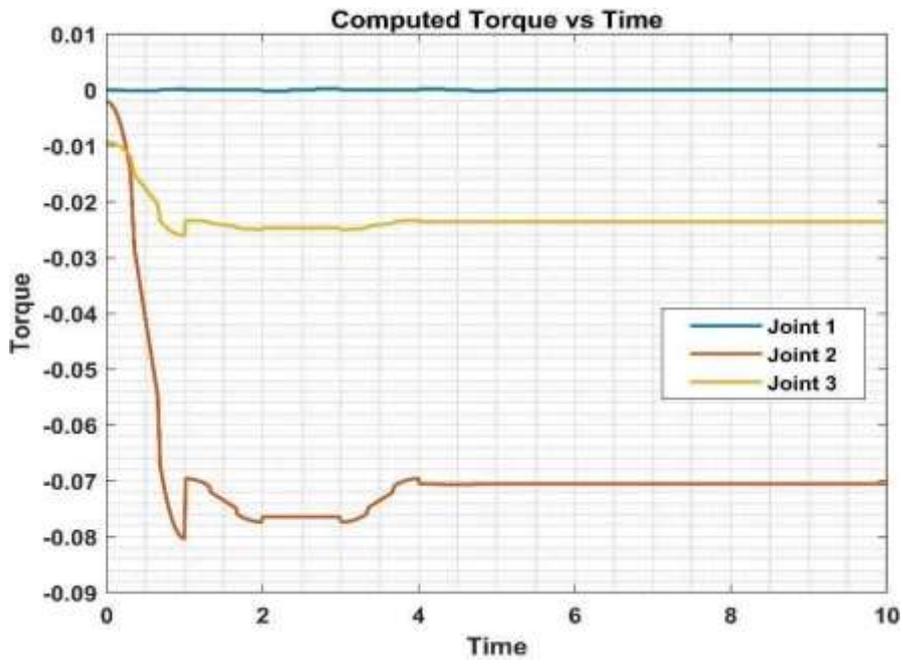
Fig 9 Computed Torque

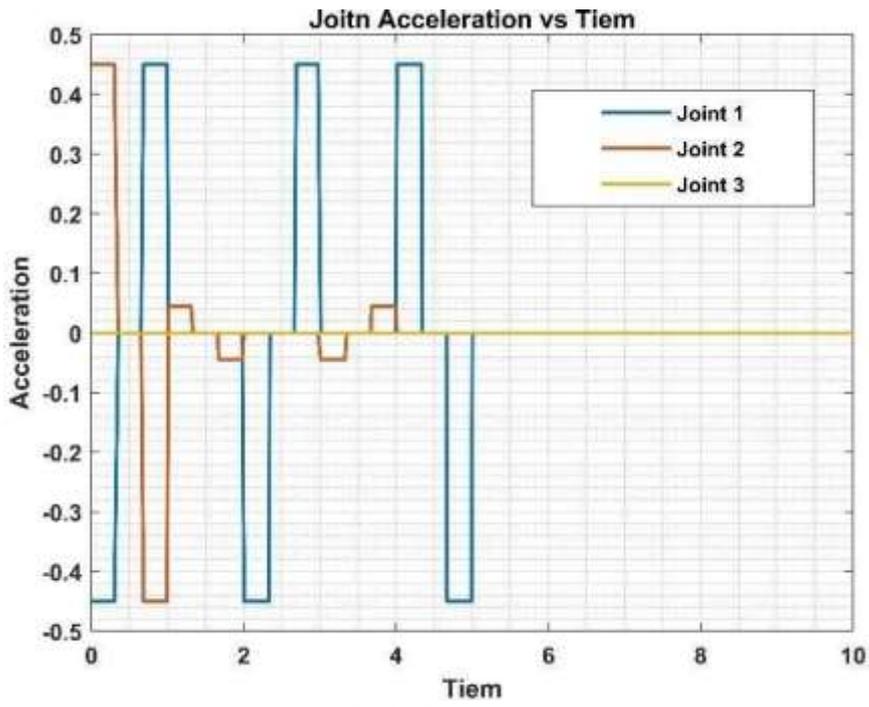
Fig 10 Joint Acceleration

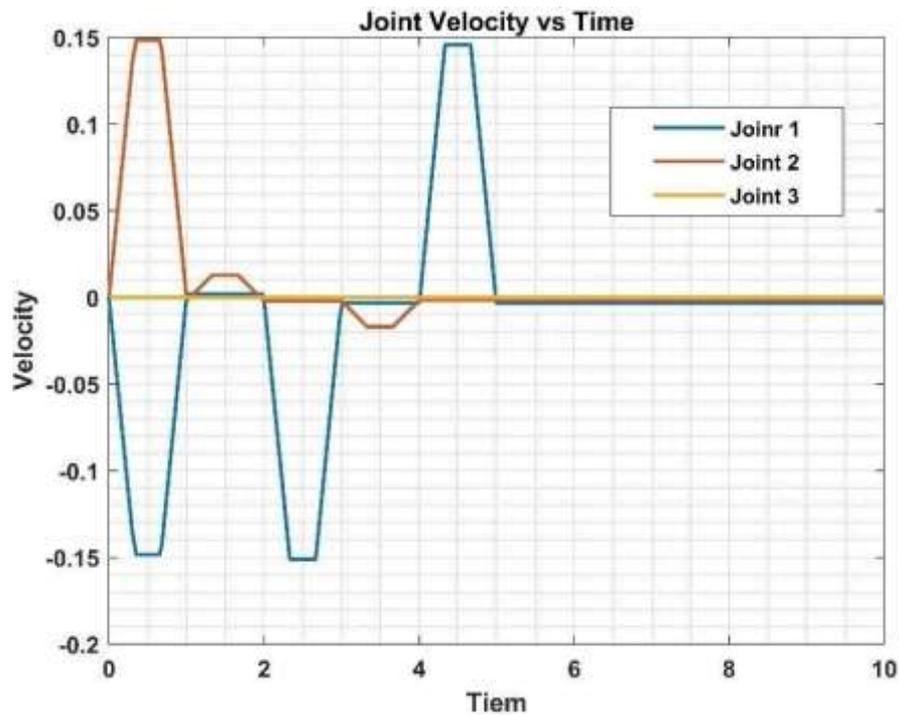
Fig 11 Joint Velocity

## 6. Conclusions

In this study, our goal was to implement a computed torque controller system for a robot manipulator with a 3 DOF RRR configuration and to evaluate the performance of the implemented controller using MATLAB. The simulation's results show that the suggested strategy is workable and offers the possibility of computed torque control for a 3DOF robot manipulator. This paper's research gives a thorough process for designing a three-DOF robotic manipulator, starting with modeling. The robotic manipulator's simulated dynamics are used in the CTC control approach. The torque of each joint is provided by the inverse dynamic block. This work gives us how to compute the torque and simulate dynamic control.

**References**

[1] R.ul Islam, J. Iqbal, S. Manzoor, A. Khalid and S. Khan, "An autonomous image-guided robotic system simulating industrial applications", 7th IEEE International Conference on System of Systems Engineering (SoSE), Italy, 2012, pp. 344-349.

[2] J. Iqbal, N.G. Tsagarakis and D.G. Caldwell, "A multi-DOF robotic exoskeleton interface for hand motion assistance", 33rd annual IEEE international conference of Engineering in Medicine and Biology Society [EMBS], Boston, US, 2011, pp. 1575-1678.

[3] J. Iqbal, A. Tahir, R.ul Islam and R.un Nabi, "Robotics for nuclear power plants – Challenges and future perspectives", IEEE International Conference on Applied Robotics for the Power Industry [CARPI], Zurich, Switzerland, Sep. 2012, pp. 151-156

[4] J.J. Craig, "Introduction to robotics: mechanics and control", 2004

[5] M.W. Spong, S. Hutchinson and M. Vidyasagar, "Robot modeling and control", New York, John Wiley & Sons, 2006

[6] Craig John H. Adaptive control of mechanical manipulators: Addison-Wesley. 1988.

[7] Choi JY, Lee JS. Adaptive iterative learning control of uncertain robotic systems. IEE Proc. Control Theory Appl. 2000; 147(2): 217-223.

[8] Slotine JJ, Li W. On the adaptive control of robot manipulators. The International Journal of Robotics Research. 1987; 6(3): 49-59.

[9] Arimoto S, Kawamura S, Miyasaki F. Bettering operation of robots by learning. Journal of Robotic Systems. 1984; 1(2):123-140.

[10] Kawamura S, Miyazaki F, Arimoto S. Realization of robot motion based on a learning method. IEEE Transactions on Systems, Man, and Cybernetics. 1988; 18(1):123-6.

[11] Li Z, Ge SS, Wang Z. Robust adaptive control of coordinated multiple mobile manipulators. Mechatronics. 2008; 18(5-6): 239-250.

[12] Craig John J. Introduction to Robotics: Mechanics and Control, Third Edition: Prentice Hall, New York. 2005.

[13] X. Shi and N.G. Fenton, "A complete and general solution to the forward kinematics problem of platform-type robotic manipulators",

IEEE International Conference on Robotics and Automation Proceedings, 1994, pp. 3055-3062

[14] R. Kumar, P. Kalra and N. Prakash, "A virtual RV-M1 robot system, robotics and computer-integrated manufacturing", Vol. 26, No. 6, 2011 pp. 994-1000

[15] S.N. Cubero, "Blind search inverse kinematics for controlling all types of serial-link robot arms", Mechatronics and Machine Vision in Practice, Springer Berlin Heidelberg, 2008

[16] K.E. Clothier and Y. Shang, "A geometric approach for robotic arm kinematics with hardware

design, electrical design and implementation", Journal of Robotics, Article ID 984823, Vol. 2010

[17] W. Khalil, "Dynamic modeling of robots using newton-euler formulation", Informatics in Control, Automation and Robotics, Springer, Berlin Heidelberg, 2011, pp. 3-20.

[18] M.P. Groover, M. Weiss and R.N. Nagel, "Industrial Robotics: Technology, Programming and Application", McGraw-Hill Higher Education, 1986

[19] Q. Li, A.N. Poo and M. Ang, "An enhanced computed-torque control scheme for robot manipulators with a neuro-compensator." Systems, Man and Cybernetics, 1995. IEEE International Conference on Intelligent Systems for the 21st Century, Vol. 1, 1995, pp. 56-60

[20] L. Reznik, "Fuzzy controllers handbook: how to design them, how they work", Access Online via Elsevier, 1997

[21] S. Soltani and F. Piltan, "Design artificial nonlinear controller based on computed torque like controller with tunable gain", World Applied Science Journal 14, No. 9, 2011, pp. 1306- 1312